\documentclass{article}
\usepackage[top=1in, bottom=1in, left=1in, right=1in]{geometry}
\usepackage{graphicx} % Required for inserting images
\usepackage[utf8]{inputenc}
\usepackage{amsmath}
\usepackage{graphicx}
\usepackage{sidecap}
\usepackage[ruled,vlined]{algorithm2e}
\usepackage{caption}
\newcommand*{\putunder}[2]{%
  {\mathop{#1}_{\textstyle #2}}%
}

\newcommand{\pluseq}{\mathrel{+}=}

\usepackage{outlines}
\usepackage{enumitem}
\setenumerate[1]{label=\Roman*.}
\setenumerate[2]{label=\Alph*.}
\setenumerate[3]{label=\roman*.}
\setenumerate[4]{label=\alph*.}

\usepackage{cite}
\usepackage{nameref,hyperref}

\title{Motif distribution and function of sparse deep neural networks}
\author{Olivia Zahn$^*$, Thomas Daniel$^{**}$, J. Nathan Kutz$^{\dag,\ddag}$\\[.2in]
$^*$Department of Physics, University of Washington, Seattle WA\\
$^{**}$Department of Biology, University of Washington, Seattle WA\\
$^\dag$Department of Applied Mathematics, University of Washington, Seattle WA\\
$^\ddag$Department of Electrical and Computer Engineering, University of Washington, Seattle WA\\}

\begin{document}

\maketitle

\section{Abstract}
We characterize the connectivity structure of feed-forward, deep neural networks (DNNs) using network motif theory. 
To address whether a particular motif distribution is characteristic of the training task, or function of the DNN, we compare the connectivity structure of 350 DNNs trained to simulate a bio-mechanical flight control system with different randomly initialized parameters. 
We develop and implement algorithms for counting 2\textsuperscript{nd}- and 3\textsuperscript{rd}-order motifs and calculate their significance using their Z-score. The DNNs are trained to solve the inverse problem of the flight dynamics model in \cite{10.1093/iob/obac039} (i.e., predict the controls necessary for controlled flight from the initial and final state-space inputs) and are sparsified through an iterative pruning and retraining algorithm \cite{zahn2022pruning}. 
We show that, despite random initialization of network parameters, enforced sparsity causes DNNs to converge to similar connectivity patterns as characterized by their motif distributions.
The results suggest how neural network function can be encoded in motif distributions, suggesting a variety of experiments for informing function and control.

\section{Introduction}
%Network science, computation, neural networks
Complex networks are prevalent in nature, technology, and in mathematical and computational modeling. 
Unlike random and lattice networks, complex networks are characterized by a non-trivial topology which enables complex collective dynamics.
The study of complex networks spans disciplines from discrete mathematics to the social sciences, but it has historically focused on natural, physical, and real-world networks. 
For example, the spread of disease is affected by the connectivity and organizing principles of societal networks \cite{pastor2015epidemic}. 
In animal neural systems, network structures and specialized synaptic pathways evolved for specific behaviors \cite{ebbesson1984evolution} or encode critical function \cite{kunert2017spatiotemporal,kutz2019neurosensory}.
% Why look at topology in complex networks...
While it is generally accepted that the behavior of a complex network is intrinsically tied to its structure and function, little is known about how (if at all) local connectivity patterns (i.e. network topology) affect the overall behavior or functionality of a network. \cite{strogatz2001exploring, newman2003structure, albert2002statistical}.
We show here that the connectivity structure of sparse deep neural networks can be characterized through its statistically significant sub-graphs, or network motifs, in order to encode function and dynamics.

% Two advances in complex networks...
Two well known topological properties of complex networks are the small-world property and the scale-free property. 
%The small-world property...
In small-world networks, a node is connected to most other nodes by way of neighboring connections \cite{watts1998collective}. 
Concretely, a small-world network is one in which the average distance between two nodes scales with the logarithm of the number of nodes in the network. 
This property allows almost all nodes in a sparsely connected network to communicate with all others in the network. 
%Example of small-wolrd 
The gene co-expression network in yeast, \textit{Saccharomyces cerevisiae}, exhibits both the small-world and scale-free property of complex networks \cite{van2004yeast}.
%The scale-free property... 
A scale-free network is one where the distribution of nodal degrees (i.e., the number of connections a given node has) follows a power law \cite{barabasi1999emergence}.
The scale-free property is exhibited in many different networks, from links between web pages to citations on publications. 
%Final sentence 
Both of these properties support the assumption that complex networks are not randomly connected, but are instead structured to enable specific dynamics, tasks, function and behavior. 

%Network motifs
The connectivity of complex networks can also be characterized by the makeup of the sub-graphs within the network. 
A network motif is a sub-graph within a larger network that occurs significantly more than it would occur in an equivalent randomly connected network \cite{alon2007network}.
Network motifs have been discovered in natural networks ranging from gene transcription to ecological networks \cite{alon2007network, milo2002network, stone2019network}. 
In some contexts, individual network motifs have clear, interpretable functions. 
For example 2\textsuperscript{nd}-order chain sub-graphs (three nodes chained together in a feed-forward manner), occur with very high significance in food-chain networks, representing the hierarchical relationship between predators and prey.
Relating the function of individual sub-graphs to the overall dynamics of a complex network is more difficult.
In \cite{hu2018feedback}, the authors relate the dynamics of a network to the statistics of its network motifs. 
The authors constrain their study to linear, time-invariant networks and derive the network transfer function (i.e., function that transforms the time-dependent input to the time-dependent output) in terms of motif cumulants (simple statistics of network motifs).  
They apply the method to several example real-world networks (power grids and \textit{C. elegans} neuronal networks) and show that a few low-order motifs are needed to model the network transfer function.
Relating the distribution of motifs to broader signal processing and functionality is of continued interest to the dynamics of networks of neurons \cite{trousdale2012impact,hu2013motif,ocker2017linking}.

%Bridge over to deep neural networks
Meanwhile, in the fields of computer science and engineering, computational models for complex tasks such as human language \cite{openai2023gpt4} and high-dimensional non-linear fluid flow \cite{Lusch_2018, Bieker_2020} have seen great success thanks primarily to advances in machine learning and specifically deep neural networks (DNNs).
%What are DNNs
%DNNs are a class of machine learning models with a network structure that was originally inspired by Hubel and Wiesel's formative study of the visual cortex \cite{hubel1962receptive, shatz1978ocular}.
%The networked structure of DNNs allows for complex interactions between input variables and make them powerful tools for modeling many tasks.
%The problem
%However, to model complex systems, DNNs often need to be extremely large. 
%Large language models 
%Present-day large language models (LLMs), DNNs that successfully simulate human languages, require hundreds of billions of parameters \cite{openai2023gpt4}.
%Such models are computationally expensive and, due to their complexity, scientists and engineers struggle to understand their decision boundaries. 
%This feature allows for the expressivity necessary to model complex systems, but is an unacceptable trade-off in fields where explainable models are absolutely critical, such as in medical diagnosis \cite{}.  
%One approach to elucidate DNNs, as well as identify excess and redundant parameters, is to approach them as complex networks, and use the tools of the field to study them. 
%Researchers have approached DNN topology from a few different angles. 
DNNs are complex networks that can be characterized by various statistical and/or topological approaches.
In \cite{naitzat2020topology}, the authors found that the topological complexity of a binary classification data set is reduced as the data passed through the network. 
In another study, researchers generate bio-instantiated recurrent neural networks, novel DNN architectures that are built from empirical data on animal neural networks \cite{goulas2021bio}.
%Conclusion and introduce/bridge to sparsity
The challenge of studying the connectivity structure of a DNN is made difficult by the sheer number of parameters in the network. 
Consequently, DNN sparsification is a helpful tool in the effort to uncover the relationship between DNN connectivity patterns and their function. 

%Bridge to sparsity & intro
There are several different methods for reducing the number of parameters in a trained DNN, including by initially defining a sparse architecture or through regularization. 
One popular method is neural network pruning, which involves the systematic removal of parameters from a trained DNN. 
Pruning was first introduced in \cite{lecun1989optimal}, in which the authors show that sparsifying a DNN via pruning improves generalization and efficiency.
Pruning is inspired by a biological process called \textit{synaptic pruning}; the elimination of synaptic connections during development. 
Synaptic pruning plays a crucial role in the refinement of neural pathways and contributes, in part, to efficient motor learning and cognitive function \cite{CRAIK2006131, morizawa2022synaptic}. 
In machine learning, several works have shown that pruning can reduce the number of parameters in a trained DNN by as much as 93\% \cite{han2015learning, zahn2022pruning}. 
Pruning has also been used to discover sub-networks that, when trained in isolation, will achieve comparable performance to the fully-connected network \cite{frankle2018lottery}.
And in sensor placement applications, pruning can evaluate optimal sensor locations \cite{williams2022data}. 
Sparsification via pruning is a simple method to reduce the number of network parameters and elucidate relevant and necessary connections for DNN performance. 

%Introduce the work done in this paper - Contributions
Here, we seek to answer the question: (i) are trained DNNs composed of a set of statistically significant sub-graphs or (ii) are they more or less randomly connected? 
We use the distribution of network motifs to characterize the connectivity structure of sparse DNNs trained to simulate a bio-mechanical flight control system \cite{zahn2022pruning}.
To address whether a particular motif makeup is characteristic of the training task, we compare the results across 350 DNNs trained to model the same system, but with different randomly initialized DNN parameters.
We develop and implement algorithms for counting 2\textsuperscript{nd}- and 3\textsuperscript{rd}-order motifs in feed-forward, sparse deep neural networks and calculate their significance using their z-score. 
The DNNs are trained to model the inverse of the flight dynamics model in \cite{10.1093/iob/obac039} (i.e., predict the controls necessary for controlled flight from the initial and final state-space inputs). 
We sparsify the DNNs through an iterative pruning and retraining algorithm \cite{zahn2022pruning}. 
This work shows that, despite random initialization of network parameters, enforced sparsity will cause DNNs to converge to similar connectivity patterns as characterized their network motif landscape. 
Concretely relating the connectivity patterns to the function of a trained DNN is beyond the scope of this work. 
However, we proposed several ideas for future experiments in the Discussion of this manuscript. 

\begin{figure}[t]
    \centering
    \includegraphics[width=\linewidth]{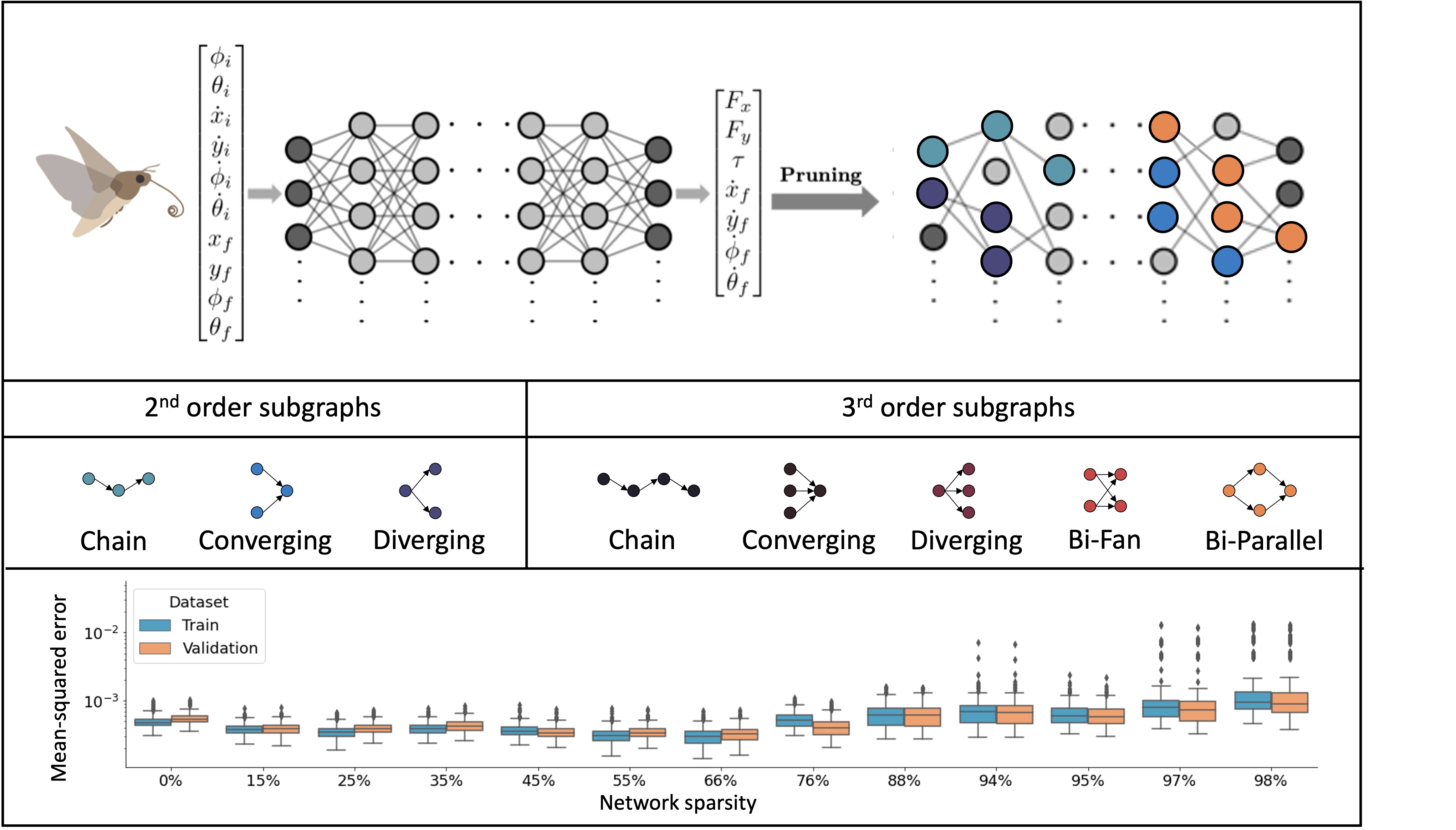}
    \caption{Top: A densely connected DNN is trained to predict the control variables for the task of insect hovering. Initial and final state-space variables are used as inputs to the network. The trained network is pruned to maximal sparsity with little decrease in performance. \hspace{5pt} Middle: A subset of 2\textsuperscript{nd}- and 3\textsuperscript{rd}-order network sub-graphs that can exist in a feed-forward DNN. \hspace{5pt} Bottom: Training and validation loss over 350 networks pruned to different sparsity levels.}
    \label{fig:fig1}
\end{figure}

\section{Methods}
The DNNs used in this study were trained to model the insect flight dynamics model described in \cite{10.1093/iob/obac039, zahn2022pruning}. 
The methods for training and pruning closely follow the procedure described in \cite{zahn2022pruning}. 
The training and pruning procedure is summarized here and differences between the two studies will be highlighted. 
All code associated with the simulations, training and pruning DNNs, and network motif analysis is available on Github \cite{MothMotifs}. %Github link.
The network characterization is performed on a specific model given that the control objectives, or the ground truth objective, is known.  Thus the results can be validated against an interpretable and well-studied model.  The algorithms developed, however, can be more broadly applied to complex networks in general with the goal of extracting insight into the underlying dynamics.

\subsection{Network training data}
The insect flight simulation uses an inertial dynamics model developed in Bustamante et al., 2022 \cite{10.1093/iob/obac039} and is inspired by the flight mechanics of the hawkmoth, \textit{Manduca sexta}.
The insect is modeled by two conjoined ellipses representing the head-thorax and abdomen body segments. 
The dynamics of the insect is constrained to two dimensions ($x$ and $y$) and is controlled by three control variables, $F$, the average force applied by the wings, $\alpha$ the direction of force applied, and $\tau$, the abdominal torque exerted about the pin joint connecting the two body segment masses.
The system's state-space is described by four parameters ($x$: horizontal position, $y$: vertical position, $\theta$: head-thorax angle, and $\phi$: abdomen angle), as well as the respective state-space derivatives ($\dot{x}$: horizontal velocity, $\dot{y}$: vertical velocity, $\dot{\theta}$: head-thorax angular velocity, and $\dot{\phi}$: abdomen angular velocity).
The model described in \cite{10.1093/iob/obac039} is a forward model, where ordinary differential equations are used to predict the insect's final position and velocities from the initial state-space variables. 
More details about the moth model can be found in \cite{10.1093/iob/obac039, zahn2022pruning}.

\subsection{Neural network training and pruning}
The above described model is used to generate data to train the DNNs studied in this work. 
All simulated trajectories were started from the origin and initial state-space and control variables were randomly sampled. 
The training dataset is comprised of 10 million simulated trajectories and the test set contains an additional 5 million trajectories.

The deep, fully-connected neural network has ten input variables ($\dot{x}_i$, $\dot{y}_i$, $\phi_i$, $\theta_i$, $\dot{\phi}_i$, $\dot{\theta}_i$, $x_f$, $y_f$, $\phi_f$, $\theta_f$) and seven output variables ($F_x$, $F_y$, $\tau$, $\dot{x}_f$, $\dot{y}_f$, $\dot{\phi}_f$, $\dot{\theta}_f$). 
Prior to pruning, all of the networks are initialized with a deep, feed-forward structure, with four hidden layers with 400, 400, 400, and 16 nodes.
The inverse tangent activation function is used in every layer to introduce non-linearity to the model. 

This is a multi-output regression model, so during training, the uniformly-weighted average of the mean squared error across the outputs was used as a loss function. 
We used the Jax deep learning library to parallelize and consequently speed up training of the models. 
The Adam optimizer and a batch size of 128 samples were likewise chosen to decrease training time.
No regularization techniques (such as weight regularization or dropout) were used, but early stopping (with a minimum validation loss delta of 0.01 and a patience of 1000 batches) was used to halt training at model convergence. 

%Training and pruning in parallel
The goal of this study is to compare the connectivity structure across many DNNs. To speed up the training and pruning of the DNNs, we parallelized the training and pruning of 350 networks. This was achieved by optimizing over the performance of all networks collective error (i.e., loss was minimized over all 350 networks). After the networks are trained to convergence, a round of neural network pruning reduces the total number of weights by some fixed and predetermined amount (e.g. $15\%$, $25\%$, etc.). The result of the experiment are snapshots of 350 networks pruned at varying levels of sparsity. 

\subsection{Post-pruning}
After the networks are trained and pruned as described above (and in detail in \cite{zahn2022pruning}), there still exist some weights that are obsolete. 
This occurs when an upstream weight is removed such that all downstream nodes no longer have an input. 
Since the pruning algorithm utilized here does not prevent such prunes, these weights are retroactively removed. 
The precise removal of these connections is described in detail in the Supplementary Material. 

\subsection{Motif Significance}
We used network motifs to access the similarity between trained DNNs connectivity structures. 
Finding a network motif of a larger network is a two step process. 
First, the total number of a desired sub-graph must be counted. 
That number is then used to calculate the significance of the the sub-graph. 
Here we use the z-score to determine the significance of a given sub-graph. 

\subsubsection{Sub-graph counting algorithms}
%Address choice to not use motif finding softwares 
There are many open source network motif counting software code bases available \cite{kashtan2004efficient, wernicke2006fanmod, meira2014acc}. 
However, these code bases are usually built for larger, more complex networks and can be slow to use. 
As such, we develop our own counting algorithms specifically tailored to the task of exactly counting sub-graphs in feed-forward, sparse DNNs. 
These algorithms take advantage of the connectivity matrices of the feed-forward DNNs and utilize simple matrix operations and combinatorics to quickly calculate the number of a given sub-graph. 

The counting algorithms can all be found in the Supplementary Material of this manuscript. 
Every counting algorithm describes the counting process for the total number of occurrences of the specified sub-graph in a single network. 
We will walk through one example here, as all of the counting algorithms follow the same general pattern. 
In the following algorithm, every pruned network is considered a list of sparse, binary matrices or masks. 
The variable \textit{mask list} refers to the list of sparse matrices associated with one pruned network. 
Likewise, \textit{mask} refers to a single sparse, binary matrix. 
Algorithm \ref{alg:2Ocon} shows the steps for counting the 2\textsuperscript{nd}-order converging sub-graph in a sparse, feed-forward DNN. 

\begin{algorithm}[h!]
\SetAlgoLined
total $= 0$\;
 \For{mask in mask list}{
    \For{column in mask}{
    $ n \leftarrow$ count number of non-zero elements\;
    \uIf{$n \geq 2$}{
    total $ \mathrel{+}=$ $n \choose 2$\;
    }
    }
 }
 \caption{2\textsuperscript{nd}-order converging sub-graph counting}
 \label{alg:2Ocon}
\end{algorithm}

The first loop in Algorithm \ref{alg:2Ocon} loops over the layers of a network. 
The second loop loops over the input nodes of the layer. 
The number of output nodes, $n$ is the number of non-zero elements connected to an input node. 
If an input node has two or more output nodes, the total number of 2\textsuperscript{nd}-order converging sub-graphs is $n \choose 2$. 
The total number of sub-graphs is tallied until all layers of the DNN have been iterated over. 

\subsubsection{Z-score calculation}
We use the Z-score to determine sub-graph significance and find network motifs. The Z-score for a given motif, $Z_m$, is defined as
\begin{equation}
    Z_{m} = \frac{N_{real}-\langle N_{random}\rangle}{\sigma_{random}}
    \label{eq:zscore}
\end{equation}
where $N_{real}$ is the total number of times the given sub-graph occurs in a sparse, DNN. 
This is found by one of the sub-graph counting algorithms described above and in the Supplemental Material. 
$N_{random}$ is the total number of times the given sub-graph occurs in an equivalently sized, but randomly connected, sparse DNN. 
This number is found by generating a randomly connected graph, with the same feed-forward structure as the trained DNNs and counting the number of sub-graphs according to the sub-graph counting algorithms. 
$Z_m$ must be calculated over the average across many randomly connected networks. 
For each of the 350 networks analyzed here (and across all sparsity levels), we generate 1000 randomly connected networks for this calculation. 
The generation of equivalently sized, but randomly connected networks is described in detail in the Supplementary Material. 
The average number of sub-graphs and the variance of the number of sub-graphs is found and used to calculate the final z-score of the motif.

\begin{figure}[h]
    \centering
    \includegraphics[width=\linewidth]{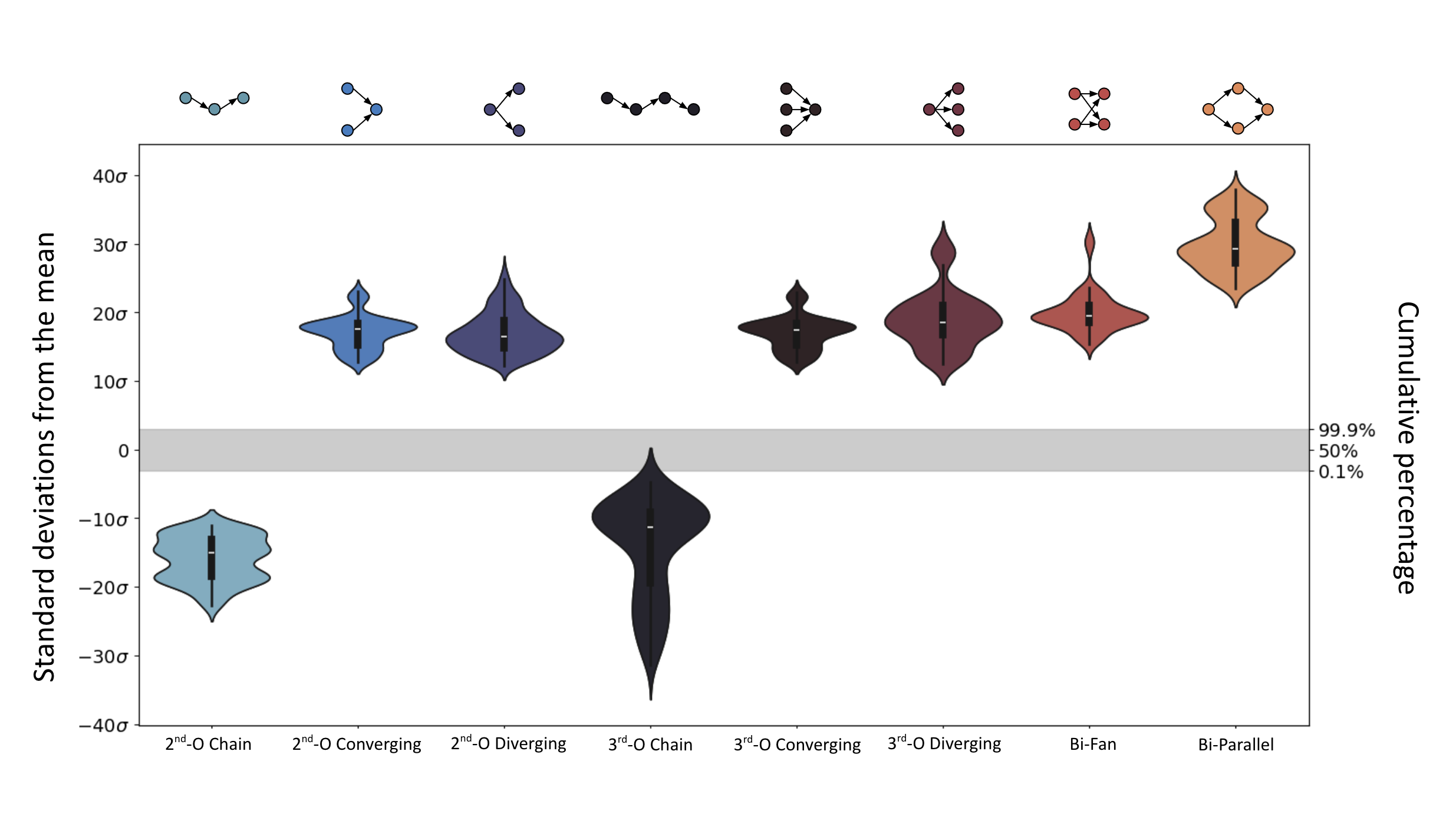}
    \caption{Distributions of z-scores across 350 DNNs pruned to 98\% sparsity. Top axis shows the motif, left axis shows the standard deviations from the mean (or z-score), and the right axis shows the cumulative percentage.}
    \label{fig:fig2}
\end{figure}

\begin{figure}[h]
    \centering
    \includegraphics[width=\linewidth]{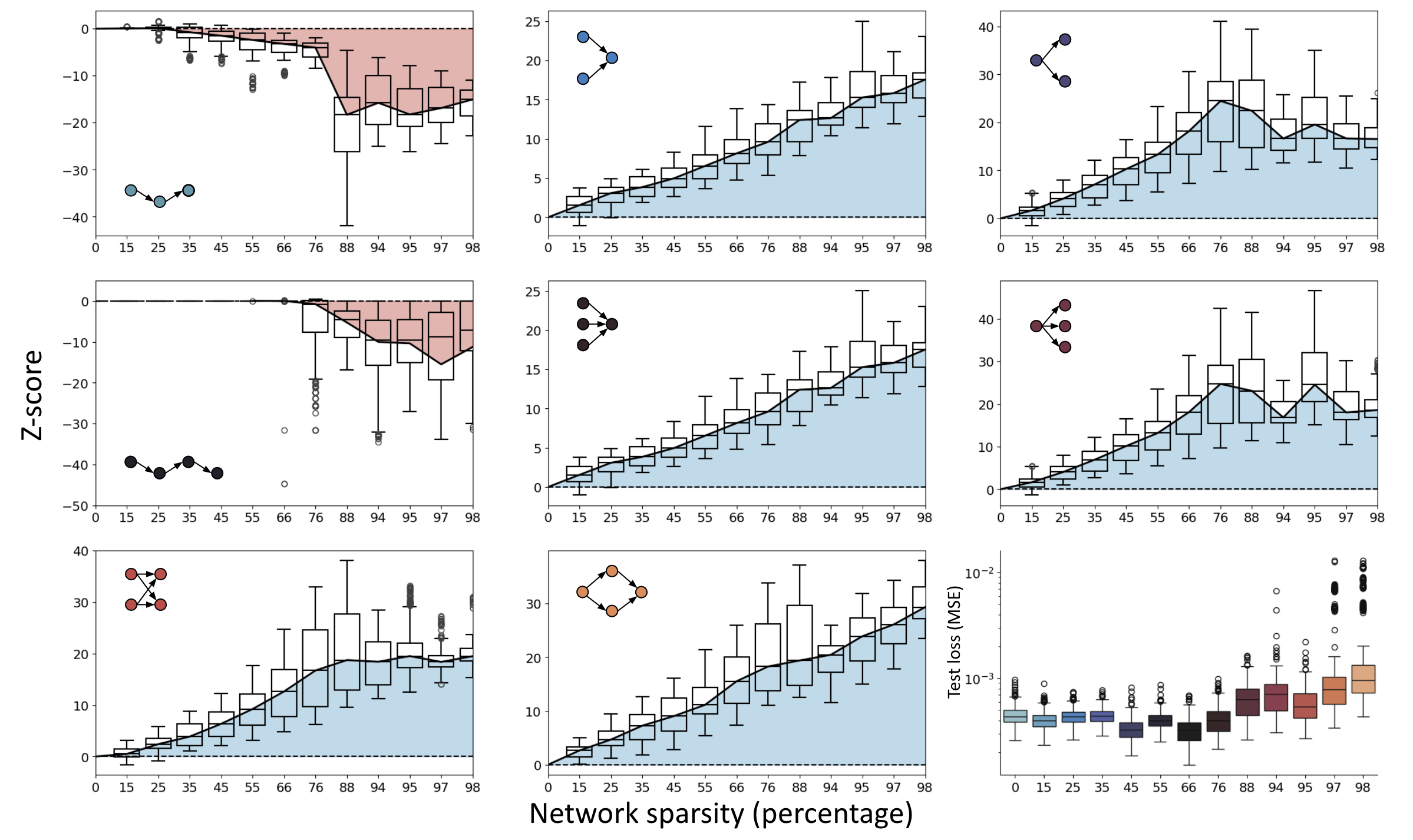}
    \caption{Z-score distributions across sparsity levels. Each panel shows how the z-score of the pictured motif changes throughout the pruning process. The bottom right panel shows the test MSE across all 350 networks at increasing levels of sparsity.}
    \label{fig:fig3}
\end{figure}

\section{Results}
The main finding of this work is that enforced sparsity during DNN training promotes network connectivity structure as characterized by network motifs. We find that there are distinct patterns in the network motif structure across networks as well as throughout the sparsification process (via pruning). 

\subsection{Patterns across networks}
Figure \ref{fig:fig2} shows the summary statistics of the network motif distributions across the 350 networks pruned to $98\%$. 
Each distribution in the violin plot shows the distribution of Z-scores for each motif over the 350 networks pruned to $98\%$.
Across the networks, distinct patterns of over- and under-representation can be seen for different network motifs. 
Furthermore, at this level of sparsity, all motifs are either highly over- or under-represented, with approximate average motif representation well above (or below) $\pm 10 \sigma$. 

\subsection{Patterns across sparsity levels}
Figure \ref{fig:fig3} shows how the network motif distribution changes as the network is pruned (according to the pruning paradigm described in the Methods section). 
Each panel shows how the representation of the motif pictured changes with the sparsity levels labeled on the x-axis (i.e., $0$ to $98\%$). 
From dense to sparse, network motifs are either over- or under-represented throughout the pruning process, except for a few outliers in the low sparsity networks (see variance bars at $15\%$ sparsity in the panels for 2\textsuperscript{nd}-order converging and diverging, 3\textsuperscript{rd}-order converging and diverging, and bi-fan motifs). 
Otherwise, network motifs are pushed to higher levels of over- or under-representation from very early in the pruning process. 

The significance of some motifs (2\textsuperscript{nd}-order converging, 3\textsuperscript{rd}-order converging, and bi-parallel) appear to increase monotonically and show no signs of leveling-off even at very high levels of sparsity. The other motifs tend to reach convergence, leveling off at some maximum or minimum level of significance. Some of these motifs (see 2\textsuperscript{nd}-order diverging) show signs of decreasing after reaching maximum significance at $76\%$ sparsity. 

\section{Discussion}
%What we did
In this study, we have developed a method to characterize the network motif distribution of sparse, deep neural networks trained to simulate insect flight control. 
Using the flight dynamics model in \cite{10.1093/iob/obac039} and the pruning paradigm developed in \cite{zahn2022pruning}, we counted the 2\textsuperscript{nd}- and 3\textsuperscript{rd}-order network motifs of 350 networks pruned to increasing levels of sparsity. 
The DNNs were trained to predict the control variables for controlled hovering from the initial and final state-space variables as inputs. 
All 350 networks were trained and pruned in parallel where the pruning process followed the sequential pattern described in \cite{zahn2022pruning}.
We developed motif counting algorithms that take advantage of the feed-forward structure of the DNNs studied in this work. 
The total number of motifs counted for each network was used to calculate the motif significance (z-score) for each network and results were compared across all 350 networks. 

%What we found
%Discussion of Figure 2
The significance of a motif is determined by comparing its number of occurrences in the network to its number of occurrences in a equivalently sized, but randomly connected network.
We constrained the random networks to a feed-forward architecture with the same number of nodes and connections per layer as the network being evaluated. 
As a network is sparsified via pruning, motifs become more positively or negatively significant (depending on the motif).
Figure \ref{fig:fig2} shows the distributions of network motif z-scores at $98\%$ sparsity. 
At high levels of sparsity, network motifs are either highly over- or under-represented.
Motifs below the midline of Figure \ref{fig:fig2} are highly under-represented. 
In other words, these motifs (2\textsuperscript{nd}-order chain and 3\textsuperscript{rd}-order chain, specifically) occur much less in the trained and pruned networks than in an equivalent random counterpart. 
The opposite is true for the the motifs above the midline. 
These motifs occur in greater numbers in the trained and pruned networks than in the random networks. 
The magnitude of the z-score for all motifs in the $98\%$ pruned networks is very large. 
Magnitude-based weight pruning (the method employed here) results in networks with highly significant sub-graph structures. 

%Discussion of Figure 3
Motif significance increases (or decreases) with network sparsity. 
Figure \ref{fig:fig3} shows how the motif significance relates to the network sparsity. 
At low levels of sparsity (in other words, nearly fully-connected layers) motifs have very low z-scores.
The low z-scores are due to the fact that there are few unique ways for the connections of a nearly fully-connected network to be organized. 
As networks are pruned and become more sparse, the over- or under-representation of a given motif becomes more pronounced. 
Some motifs become more under-represented as the network is made more sparse (i.e., 2\textsuperscript{nd}-order chain and 3\textsuperscript{rd}-order chain motifs). 
All of the other motifs become over-represented as more connections are pruned from the network. 
Furthermore, the significance levels for the 2\textsuperscript{nd}-order converging, 3\textsuperscript{rd}-order converging, and the bi-parallel motifs all monotonically increase. 
In contrast, at higher levels of sparsity, all other motifs display a leveling-off in their significance levels.

There are clear patterns when looking at different orders of the same motif type (e.g. the 2\textsuperscript{nd}-order and 3\textsuperscript{rd}-order chain motifs)
Although the 2\textsuperscript{nd}-order and 3\textsuperscript{rd}-order chain motifs motifs becomes more significantly negative over the course of pruning, a leveling-off in the z-score occurs for high levels of sparsity. 
This is most clearly seen in the uppermost left panel of Figure \ref{fig:fig3} where at around $88\%$ sparsity, the average significance of the 2\textsuperscript{nd}-order chain motif becomes constant despite increasing sparsity. 
From this result, it can be concluded that as a feed-forward network is pruned, the chain motifs become more rare, but a minimum number of chain sub-graphs are necessary for the network to perform its function. 

%Bifan and biparallel constrained by converging 
A similar leveling-off pattern can be seen in the motif significance distributions of the 2\textsuperscript{nd}-order and 3\textsuperscript{rd}-order diverging motifs and the bi-fan motif (lower-most left panel of Figure \ref{fig:fig3}). The bi-fan and bi-parallel motifs are made up of two connected 2\textsuperscript{nd}-order converging and diverging motifs. 
The significance of both the 2\textsuperscript{nd}-order diverging and bi-fan motifs levels-off at high levels of sparsity around $20 \sigma$. 
Interestingly, the 2\textsuperscript{nd}-order converging and bi-parallel motifs do not display the same leveling-off at the tested sparsity levels. 
It is possible that with further fine-grain pruning the motif z-scores for the 2\textsuperscript{nd}-order and 3\textsuperscript{rd}-order converging motifs and the bi-parallel motif would level-off. 

%Discuss supplemental figures?
All motif pairs tested in this study (i.e., converging, diverging, and chain) have similar motif significance distributions across varying levels of sparsity. 
The close similarity between the significance distributions of the 2\textsuperscript{nd}-order and 3\textsuperscript{rd}-order converging motifs, for example, are natural due to the similarity between the sub-graphs themselves. 
While the distributions of the z-scores look very similar, differences in the distributions of the components of the z-scores are more pronounced. 
These distributions can be seen in Section \ref{supmat}. 

In the motif significance distributions in Figure \ref{fig:fig3}, the raw count of each sub-graph is very large, despite high-levels of global network sparsity (see Figure \ref{fig:motif_sup1} in Section \ref{supmat}.
This results in low resolution differences in the visualization of the motif significance distributions, specifically in the panels for the 2\textsuperscript{nd}-order and 3\textsuperscript{rd}-order converging motifs. 
For example, in a given network pruned to $98\%$ sparsity, the raw count of 2\textsuperscript{nd}-order converging sub-graphs is $64,249$ and the raw count for 3\textsuperscript{rd}-order converging sub-graphs is $64,182$. 
Differences in the sub-graph count that are on the order of $10$s or $100$s are not visible in the visualizations provided here. 
However, further experimentation and a different pruning paradigm may help distinguish these motifs. 
This experiment was done with global neural network pruning. 
For a given network, pruning was halted when network performance dropped significantly which typically occurred when outputs were pruned from the network. 
More pronounced differences in the motif significance distributions may arise with further pruning of only the hidden layers (i.e., with the input and output layers frozen at $98\%$ sparsity. 

%Different initialization and different pruning algorithms 
The DNNs analyzed here were randomly initialized prior to training, but were architecturally all the same. 
Additionally, every network underwent the same pruning paradigm. 
We followed the sequential, magnitude-based pruning schedule developed in \cite{zahn2022pruning}.
While magnitude-based pruning is an effective DNN sparsification technique, it requires the assumption that the parameters that are critical to predictive performance are the ones that impart high activation to the nodes of the network. 
Therefore, high-activation nodes and connections  are used to calculate motif significance.
An interesting experiment would be to compare the motif makeup of a fully-connected network to its sparse counterpart, where the connections and nodes used in the sub-graph counting algorithms are ones with high activation. 
It is possible that a fully-connected network is  equivalent to its sparse counterpart topologically, but with the advantage of parameter redundancy, which may increase its robustness to noisy data. 
Another interesting followup experiment would be to test if different pruning techniques affect the performance and motif makeup of a sparse network. 
Random pruning and retraining can also result in a performant DNN, however the resultant network usually performs worse than underwent structured pruning \cite{blalock2020state}.
Comparing the motif makeup of a randomly pruned DNN to a magnitude-based pruned DNN may provide insight to the importance of network sub-structures to network performance.

%How could this be more complicated? Residual layers, feed-back motifs 
Finally, this work was done with feed-forward DNNs, which are the simplest example of DNNs after perceptrons. 
We limited the scope of this work to feed-forward DNNs because they are both relevant and tractable. 
Feed-forward DNNs are able to model many complex regression problems found in physics, biology, and engineering. 
Additionally, they are computationally easy to understand and their structure allowed us to quickly calculate a subset of the possible low-order network motifs. 
There are more motifs possible in feed-forward DNNs than the ones that are presented here. 
We constrained our study to the motifs that are relevant in the literature and natural networks \cite{alon2007network, zambra2020emergence}.
Furthermore, \cite{hu2018feedback} showed that low-order motifs are the most important in relation to global network behavior. 
More complex motifs are possible in more complex network architectures. 
For example, networks with feed-back (such as recurrent neural networks) contain feed-back motifs. 
The sub-graphs we highlight connect subsequent layers, but in residual networks, information can travel further downstream without passing through every feed-forward layer. 
This work demonstrates our efforts to apply a method from complex network theory to the state-of-the-art technology. 

\section{Conclusion}
The work done here is a first step in using the tools of network theory, namely network motif theory, to characterize the connectivity landscape of a DNN trained to model a bio-mechanical task. 
The work in \cite{zahn2022pruning} showed that DNNs trained to model the same system will converge to similar levels of sparsity when sparsified via pruning. 
This work is a direct successor to that study and further demonstrates the similarity between the networks studied. 
The gradient descent and backpropagation algorithms do not guarantee the randomly initialized networks to converge to the same solution even when trained on the same data.
However, we have shown here that training a DNN results in a network structure that has a characteristic network motif landscape. 

\section{Acknowledgements}
We gratefully acknowledge the support of NVIDIA Corporation with the donation of the Titan Xp GPU used for this research and Henning Lange for his valuable help in developing the JAX code.
The authors acknowledge support from the Air Force Office of Scientific Research (FA9550-19-1-0386).
JNK further acknowledges support from the National Science Foundation AI Institute in Dynamic Systems (grant number 2112085)

\bibliographystyle{IEEEtran}
\bibliography{biblio.bib}

\section{Supplementary Material}\label{supmat}
\subsection{Sub-graph counting algorithms}

\begin{figure}[h!]
  \begin{minipage}{.2\textwidth}
  \centering
    \includegraphics[scale=0.35]{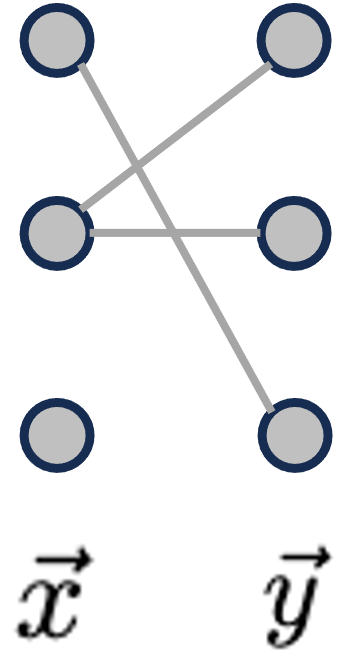}
  \end{minipage}%
    \begin{minipage}{.3\textwidth}
    \begin{equation*}
    \vec{y} = \sigma((\mathbf{W} \odot \mathbf{M})\vec{x})
    \end{equation*}
  \end{minipage}%
  \begin{minipage}{.5\textwidth}
    \begin{equation*}
    \putunder{
    \begin{bmatrix}
    0.03 & 0.02 & 0.50\\
    0.55 & 0.60 & 0.01\\
    0.01 & 0.02 & 0.01
    \end{bmatrix}}{\mathbf{W}}
    \odot
    \putunder{
    \begin{bmatrix}
    0 & 0 & 1\\
    1 & 1 & 0\\
    0 & 0 & 0
    \end{bmatrix}}{\mathbf{M}}
    \end{equation*}
  \end{minipage}%
  \caption{Left: Example of two-layer sparse network with inputs, $\vec{x}$, and outputs, $\vec{y}$. \hspace{5pt} Center: Forward pass computation where $\mathbf{W}$ represents the weight matrix $\mathbf{M}$ represents the mask matrix, and $\sigma$ represents the nonlinear activation function (bias is excluded for simplicity). \hspace{5pt} Right: Example of weight matrix $\mathbf{W}$ and mask matrix $\mathbf{M}$.}
  \label{fig:mask_ex}
\end{figure}

The following sub-graph counting algorithms take advantage of the network connectivity information contained in the sparse, binary masking layers. For example, take the sparse, two layer network depicted below. The number of connections that are still "live" in the pruned network can be ascertained entirely from the binary mask (matrix $\mathbf{M}$ in Figure \ref{fig:mask_ex}) by simply counting the number of ones in $\mathbf{M}$. Moreover, the number of more complex sub-graphs can be calculated using the binary masks of subsequent layers. 

\begin{figure}[h!]
\begin{minipage}{.25\textwidth}
        \centering
    \includegraphics[scale=0.35]{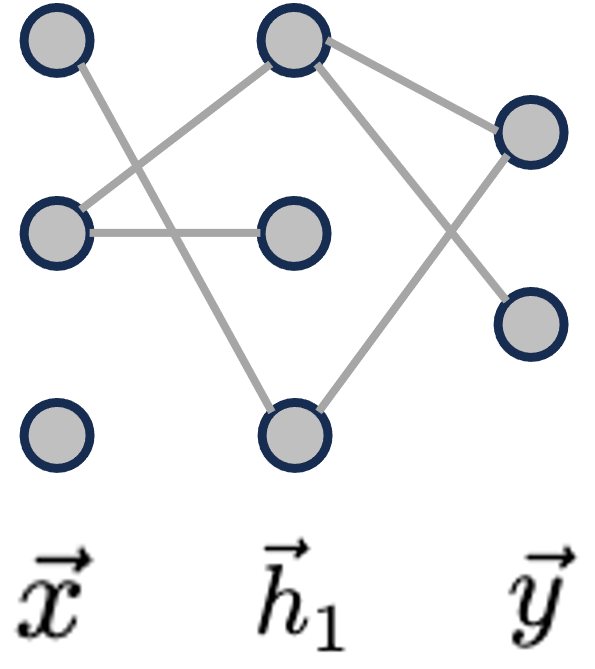}
\end{minipage}
\begin{minipage}{.25\textwidth}
    \centering
    \includegraphics[scale=0.35]{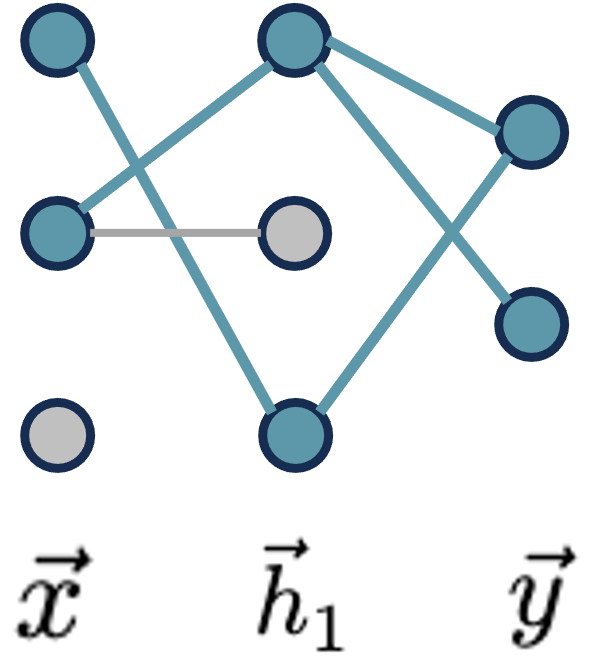}
\end{minipage}
\begin{minipage}{.5\textwidth}
\begin{equation*}
    \mathbf{M_{x,h_1}} =
    \begin{bmatrix}
        0 & 0 & 1\\
        1 & 1 & 0\\
        0 & 0 & 0
    \end{bmatrix}
   ,
    \quad 
    \mathbf{M_{h_{1}, y}} =
    \begin{bmatrix}
        1 & 1\\
        0 & 0\\
        1 & 0
    \end{bmatrix}
\end{equation*}
    
\end{minipage}

    \caption{Left: Example of sparse feed-forward network with inputs, $\vec{x}$, outputs, $\vec{y}$, and one hidden layer, $\vec{h}_1$. \hspace{5pt} Middle: Same network with second-order chain sub-graphs highlighted. \hspace{5pt} Right: Masks representing the connectivity of the network between the layers (e.g., $\mathbf{M_{x,h_1}}$ for the weights between layers $\vec{x}$ and $\vec{h}_1$).}
    \label{fig:chain_ex}
\end{figure}

For example, to find the number of second-order chain sub-graphs in the example network depicted in Figure \ref{fig:chain_ex}, we need the masks between each of the layers. The highlighted version of the network (center panel in Figure \ref{fig:chain_ex}) shows that there are three second-order chain sub-graphs in the network. The network has one hidden layer, therefore, its connectivity is entirely described by two masks $\mathbf{M_{x,h_1}}$ and $\mathbf{M_{h_{1}, y}}$. Using these two matrices, the number of second-order chain sub-graphs in the 3-layer network can be calculated by matrix multiplying $\mathbf{M_{x,h_1}}$ and $\mathbf{M_{h_{1}, y}}$ and taking the sum of the elements in the resultant product. 

\begin{align*}
\mathbf{M_{x,h_1}} \mathbf{M_{h_{1}, y}} =
    \begin{bmatrix}
        0 & 0 & 1\\
        1 & 1 & 0\\
        0 & 0 & 0
    \end{bmatrix}
    &\begin{bmatrix}
        1 & 1\\
        0 & 0\\
        1 & 0
    \end{bmatrix} =
    \begin{bmatrix}
        1 & 0\\
        1 & 1\\
        0 & 0
    \end{bmatrix} = P_{i,j}\\
    \sum_{i=1}^{m} \sum_{j=1}^{n} P_{i,j} &= 
    3
\end{align*}

To count the second-order chain sub-graphs in a n-layer feed-forward network, one has to repeat the above calculation for each consecutive pair of layers in the network and sum all of the resultant products. Algorithm \ref{alg:2Ochain} describes the second-order chain sub-graph calculation in algorithmic form. 

\begin{figure}[h!]
\begin{minipage}{.3\textwidth}
        \centering
    \includegraphics[scale=0.35]{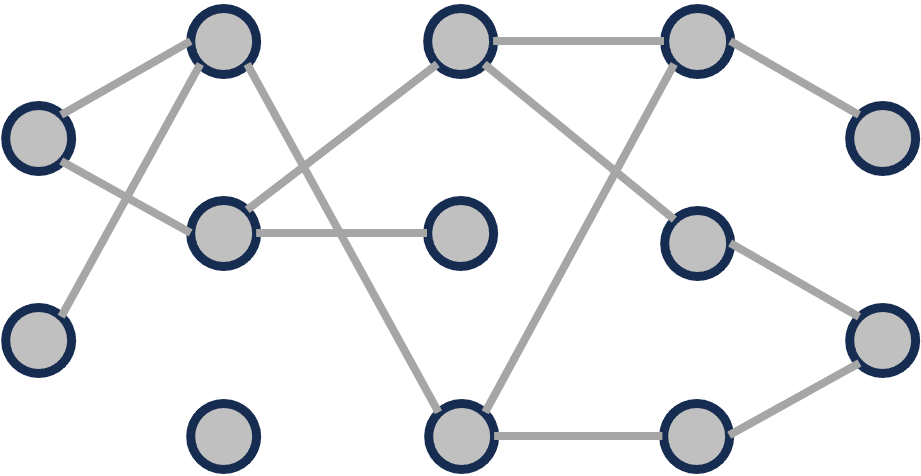}
\end{minipage}
\begin{minipage}{.6\textwidth}
\begin{equation*}
    \Longrightarrow
    \left(
    \begin{bmatrix}
        1 & 1 & 0\\
        1 & 0 & 0
    \end{bmatrix}
    ,
    \begin{bmatrix}
        0 & 0 & 1\\
        1 & 1 & 0\\
        0 & 0 & 0
    \end{bmatrix}
   ,
   \begin{bmatrix}
        1 & 1 & 0\\
        0 & 0 & 0\\
        1 & 0 & 1
    \end{bmatrix}
    ,
    \begin{bmatrix}
        1 & 0\\
        0 & 1\\
        0 & 1
    \end{bmatrix}
    \right)
\end{equation*}
    
\end{minipage}
    \caption{Example of sparse, feed-forward network converted to list of masks, or \textit{mask list} as referred to in algorithms.}
    \label{fig:nettomask}
\end{figure}

In the following algorithms, every pruned network is considered a list of sparse, binary matrices or masks. A simple example of this conversion is shown in Figure \ref{fig:nettomask}. The variable \textit{mask list} refers to the list of sparse matrices associated with one pruned network. Likewise, \textit{mask} refers to a single sparse, binary matrix. Every algorithm describes the counting process for the total number of occurrences of the specified sub-graph in a single network. 

\subsubsection{Second-order sub-graph counting algorithms}

\begin{algorithm}[H]
\SetAlgoLined
total $= 0$\;
 \For{mask in mask list}{
    product $ \leftarrow$ matrix multiply mask and subsequent mask\;
    $ n \leftarrow$ sum elements of product\;
    total $\pluseq n$

 }
 \caption{2\textsuperscript{nd}-order chain sub-graph counting}
 \label{alg:2Ochain}
\end{algorithm}

\hspace{0.5cm}

\begin{algorithm}[H]
\SetAlgoLined
total $= 0$\;
 \For{mask in mask list}{
    \For{column in mask}{
    $ n \leftarrow$ count number of non-zero elements\;
    \uIf{$n \geq 2$}{
    total $ \mathrel{+}=$ $n \choose 2$\;
    }
    }
 }
 \caption{2\textsuperscript{nd}-order converging sub-graph counting}
 \label{alg:2Oconsup}
\end{algorithm}

\hspace{0.5cm}

\begin{algorithm}[H]
\SetAlgoLined
total $= 0$\;
 \For{mask in mask list}{
    \For{row in mask}{
    $ n \leftarrow$ count number of non-zero elements\;
    \uIf{$n \geq 2$}{
    total $ \mathrel{+}=$ $n \choose 2$\;
    }
    }
 }
 \caption{2\textsuperscript{nd}-order diverging sub-graph counting}
 \label{alg:2Odiv}
\end{algorithm}

\hspace{0.5cm}

\subsubsection{Third-order sub-graph counting algorithms}
The 3\textsuperscript{rd}-order chain sub-graph counting algorithm requires the multiplication of three consecutive masks to obtain the number of 3\textsuperscript{rd}-order chains, which involves 4 nodes. 
In the following algorithm $m_i$ is used to denote mask $i$ within the network. 

\hspace{0.5cm}

\begin{algorithm}[H]
\SetAlgoLined
total $= 0$\;
 \For{mask in mask list}{
    product $ \leftarrow$ matrix multiply $m_i$, $m_{i-1}$, and $m_{i-2}$\;
    $ n \leftarrow$ sum elements of product\;
    total $\pluseq n$
}
\caption{3\textsuperscript{rd}-order chain sub-graph counting}
\label{alg:3Ochain}
\end{algorithm}

\hspace{0.5cm}

\begin{algorithm}[H]
\SetAlgoLined
total $= 0$\;
 \For{mask in mask list}{
    \For{column in mask}{
    $ n \leftarrow$ count number of non-zero elements\;
    \uIf{$n \geq 3$}{
    total $ \mathrel{+}=$ $n \choose 3$\;
    }
    }
 }
 \caption{3\textsuperscript{rd}-order converging sub-graph counting}
 \label{alg:3Ocon}
\end{algorithm}

\hspace{0.5cm}

\begin{algorithm}[H]
\SetAlgoLined
total $= 0$\;
 \For{mask in mask list}{
    \For{row in mask}{
    $ n \leftarrow$ count number of non-zero elements\;
    \uIf{$n \geq 3$}{
    total $ \mathrel{+}=$ $n \choose 3$\;
    }
    }
 }
 \caption{3\textsuperscript{rd}-order diverging sub-graph counting}
 \label{alg:3Odiv}
\end{algorithm}

\hspace{0.5cm}

\begin{algorithm}[H]
\SetAlgoLined
total $= 0$\;
 \For{mask in mask list}{
    \For{row in mask}{
    $ \mathrm{list} \leftarrow$ dot product of row with all other rows in mask\;
        \For{element in $\mathrm{list}$}{
        \uIf{element $\geq 2$}{
        total $ \mathrel{+}=$ $n \choose 2$\;
        }
        }
        
    }
    }
 \caption{Bi-fan sub-graph counting}
 \label{alg:bifan}
\end{algorithm}

\hspace{0.5cm}

\begin{algorithm}[H]
\SetAlgoLined
total $= 0$\;
 \For{mask in mask list}{
    product $ \leftarrow$ matrix multiply $m_i$, $m_{i+1}$\;
    combination matrix $ \leftarrow {\mathrm{product} \choose 2}$\;
    $n \leftarrow$ sum elements of combination matrix\;
    total $ \mathrel{+}= n$ 
 }
 \caption{Bi-parallel sub-graph counting}
 \label{alg:bipar}
\end{algorithm}

%\subsection{Z-score calculation}

\begin{figure}
    \centering
    \includegraphics[width=\linewidth]{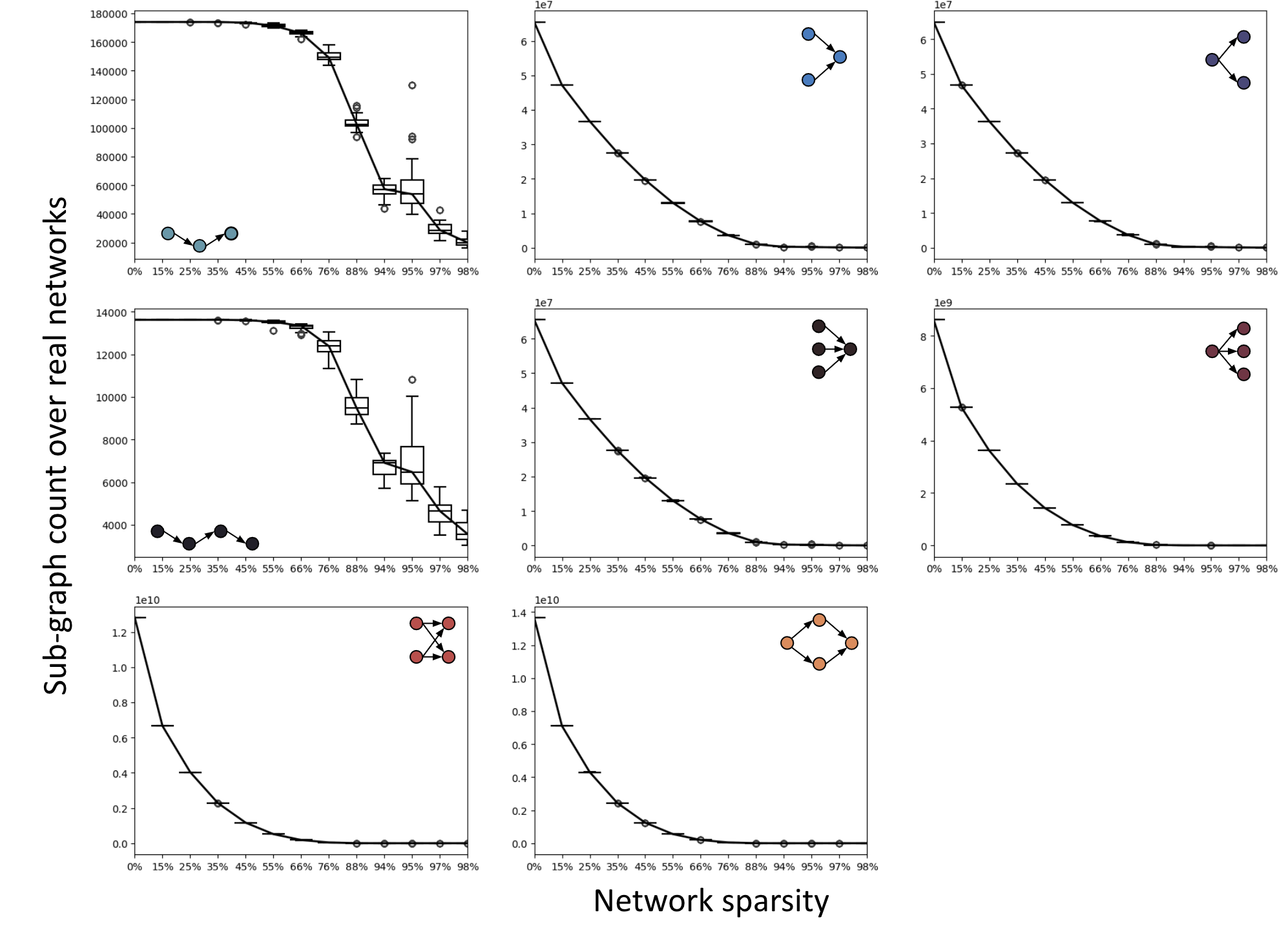}
    \caption{Total sub-graph count in 350 sparse networks across sparsity levels. Each panel gives results for each sub-graph type. Used in the z-score calculation to produce results in Figure \ref{fig:fig3}.}
    \label{fig:motif_sup1}
\end{figure}

\begin{figure}
    \centering
    \includegraphics[width=\linewidth]{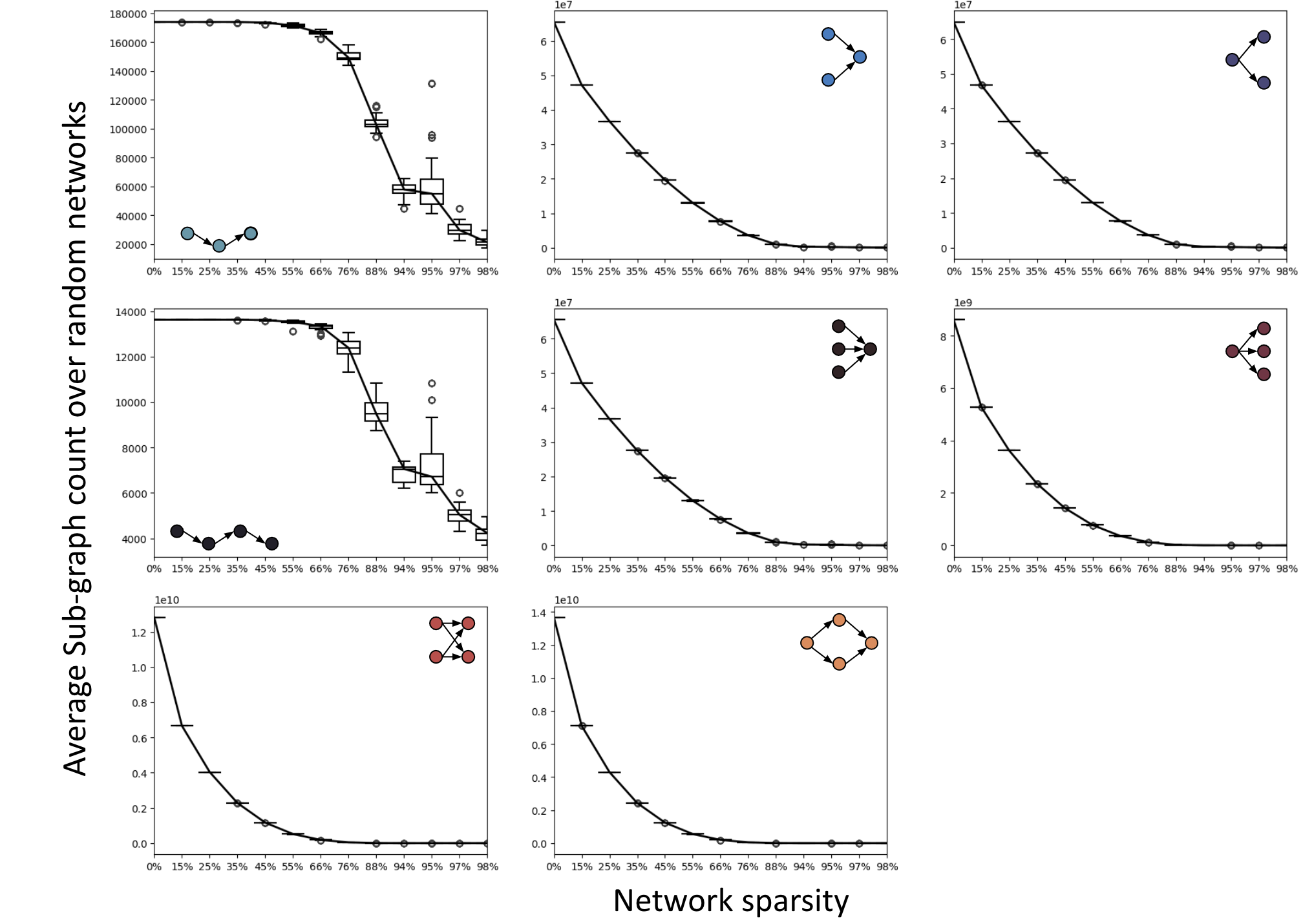}
    \caption{Average sub-graph count across 1000 random networks generated for each of the 350 sparse networks across sparsity levels. Each panel gives results for each sub-graph type. Used in the z-score calculation to produce results in Figure \ref{fig:fig3}.}
    \label{fig:motif_sup2}
\end{figure}

\begin{figure}
    \centering
    \includegraphics[width=\linewidth]{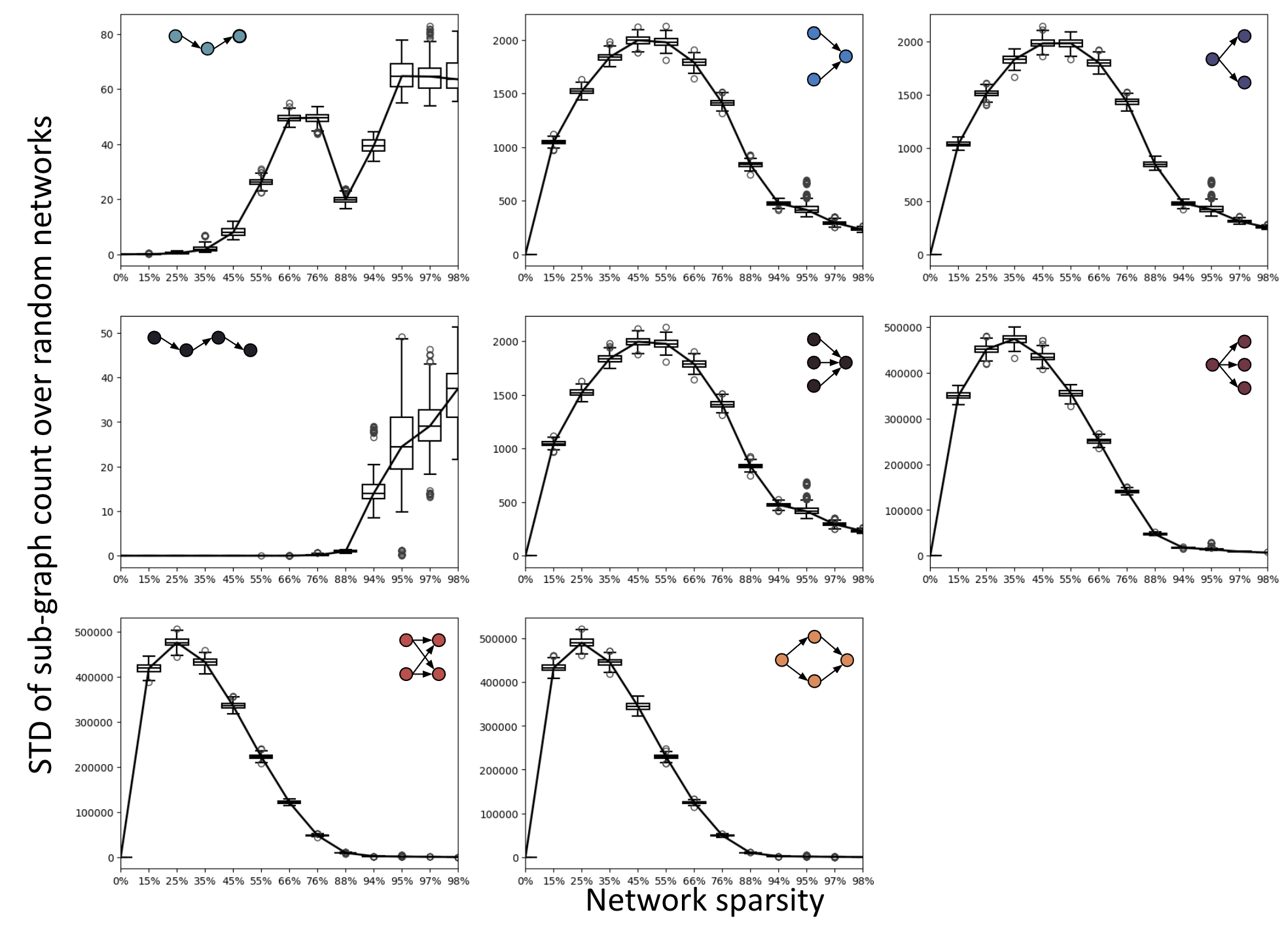}
    \caption{Standard deviation of the sub-graph count across 1000 random networks generated for each of the 350 sparse networks across sparsity levels. Each panel gives results for each sub-graph type. Used in the z-score calculation to produce results in Figure \ref{fig:fig3}.}
    \label{fig:motif_sup3}
\end{figure}

\end{document}